\documentclass[runningheads]{llncs}
\usepackage[T1]{fontenc}
\usepackage{graphicx,verbatim}
\usepackage{hyperref}
\usepackage{color}

\usepackage{booktabs}
\usepackage{tabularx} 
\newcolumntype{C}{>{\centering\arraybackslash}X}
\newcolumntype{R}{>{\raggedleft\arraybackslash}X}

\usepackage{pifont}
\newcommand{\cmark}{\ding{51}}%
\newcommand{\xmark}{\ding{55}}%

\usepackage[table]{xcolor}
\newcommand{\gmidrule}[0]{\arrayrulecolor{black!30}\midrule\arrayrulecolor{black}}

\newcommand{\posv}[1]{{\color{blue!80!black}#1}}
\newcommand{\negv}[1]{{\color{red!60!black}#1}}
\newcommand{\ind}[1]{{\color{brown!70!black}#1}}

\newcommand*{\simsym}{\mathord\sim}
\begin{document}
\title{One Sequence to Segment Them All: Efficient Data Augmentation for CT and MRI Cross-Domain 3D Spine Segmentation}
\titlerunning{One Sequence to Segment Them All}
%
\author{Nathan Molinier*\inst{1,2}\orcidID{0000-0002-8312-8952} \and
Hendrik Möller*\inst{3,4}\orcidID{0009-0001-1978-5894} \and
Thomas Dagonneau\inst{1,2}\orcidID{0009-0003-4576-4792} \and
Anna Curto-Vilalta\inst{4,5}\orcidID{0000-0002-6625-3639} \and
Robert Graf\inst{3,4}\orcidID{0000-0001-6656-3680} \and
Matan Atad\inst{3,4}\orcidID{0000-0001-6952-517X} \and
Daniel Rueckert\inst{4,6}\orcidID{0000-0002-5683-5889} \and
Jan S. Kirschke\inst{3}\orcidID{0000-0002-7557-0003} \and
Julien Cohen-Adad\inst{1,2}\orcidID{0000-0003-3662-9532}}
\authorrunning{Molinier and Möller et al.}
%
\institute{NeuroPoly Lab, Institute of Biomedical Engineering, Polytechnique Montreal \and 
Mila -- Quebec AI Institute \and
Department for Interventional and Diagnostic Neuroradiology, TUM University Hospital \and
Chair for AI in Healthcare and Medicine, Technical University of Munich (TUM)\and 
Department of Orthopedics and Sports Orthopedics, TUM University Hospital\and
Department of Computing, Imperial College London}


  
\maketitle              
\begin{abstract}

Deep learning–based medical image segmentation is increasingly used to support clinical diagnosis and develop new treatment strategies. However, model performance remains limited by the scarcity of high-quality annotated data and insufficient generalization across imaging protocols. This limitation is particularly evident in MRI and CT, where models are typically trained on a single acquisition sequence and exhibit reduced robustness when applied to unseen sequences or contrasts. Although data augmentation is widely used to improve general robustness on medical images, its impact on cross-modality generalization has not been quantitatively explored.
In this work, we study a targeted set of data augmentation techniques designed to improve cross-modality transfer. We train three spine segmentation models, each on a single-modality/sequence dataset, and evaluate them across seven out-of-distribution datasets (spanning CT and MRI), reflecting a realistic single-sequence training and multi-sequence/contrast/modality deployment scenario. Our results demonstrate substantial performance gains on unseen domains (average Dice gain of $155\%$) while preserving in-domain accuracy (average Dice decrease of $0.008\%$), including effective transfer between CT and MRI.
To mitigate the computational cost typically associated with strong data augmentation, we implement GPU-optimized augmentations that maintain, and even improve, training efficiency by approximately $\simsym10\%$. We release our approach as an open-source toolbox (\href{https://anonymous.4open.science/r/blinded-repository/README.md}{blinded hyperlink}), enabling seamless integration into commonly used frameworks such as nnUNet and MONAI. These augmentations significantly enhance robustness to heterogeneous clinical imaging scenarios without compromising training speed.


\end{abstract}
\section{Introduction}

Medical image segmentation is essential for computer-aided diagnosis and treatment development, enabling quantitative analysis and disease monitoring \cite{menze2014multimodal}. However, manual segmentation is costly, time-consuming, and prone to variability, motivating the use of automatic deep learning–based methods \cite{rayed2024deep}.

Despite strong performance, most models are tailored to specific imaging modalities/contrasts and often fail when applied to data from different modalities (CT, MRI), scanners (vendor, field strengths, software version) or protocols. This issue is particularly pronounced in spine segmentation, where methods are typically modality-specific \cite{xie2025deep}. Although CT and MRI provide complementary information, the underlying anatomy remains consistent, with domain shifts mainly arising from contrast and intensity differences.

Data augmentation is widely used to improve generalization by increasing data variability. Numerous techniques have been proposed, from basic geometric and intensity transformations to synthetic data generation \cite{billot2023synthseg,graf2024modeling,hantze2025mri,ouyang2022causality,shin2018medical,xu2020robust}. However, results are highly dataset- and task-dependent, and many studies conflate augmentation effects with architectural or dataset changes \cite{chlap2021review,kebaili2023deep,schwonberg2025domain}. Furthermore, controlled analyses in 3D segmentation are scarce, with most work focused on 2D settings \cite{goceri2023medical,schwonberg2023augmentation}. Finally, some methods proposed in the literature are difficult to adopt in practice, as they require auxiliary training, specialized pipelines, and can be computationally expensive \cite{billot2023synthseg,graf2023denoising}.

In this work, we investigate a combination of data augmentations for 3D spine segmentation using nnUNet as a standardized baseline \cite{isensee2021nnu}. Models are trained on single-sequence datasets and evaluated on out-of-distribution data spanning multiple modalities (CT, MRI), contrasts (e.g. T1w, T2w, Dixon in-phase) and sequences (e.g., resolution, field-of-view). 
Our contributions are as follows:
\begin{itemize}
    \item We propose a set of data augmentation techniques for domain generalization across CT and MRI sequences for spinal segmentation.
    \item While traditional frameworks rely on CPU for data augmentation, we present a software solution that performs data augmentation on GPU for nnUNet \cite{isensee2021nnu} and MONAI \cite{cardoso2022monai} frameworks.
\end{itemize}
\section{Methodology}

\subsection{Dataset}

\begin{figure}[!b]
    \centering
    \includegraphics[width=0.99\linewidth]{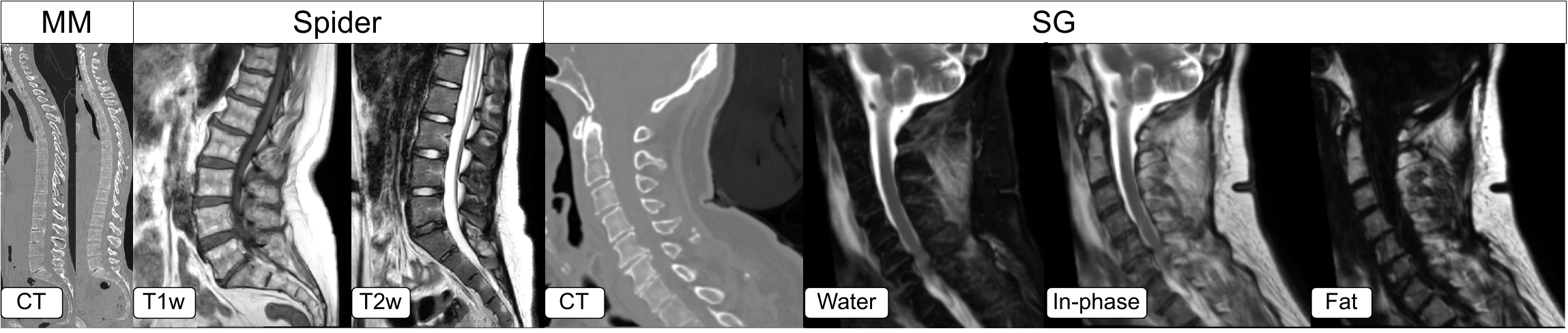}
    \caption{Example images from the different datasets. In order left to right: MM (CT), Spider (MRI T1w, T2w) \cite{van2024spider} and SG (CT, MRI Dixon water, in-phase, and fat).}
    \label{fig:dataset-show}
\end{figure}
\begin{table}[!t]
    \centering
    \caption{The demographics of the utilized datasets.}
    \label{tab:dataset_desc}
    {\fontsize{8}{9}\selectfont
    \begin{tabularx}{0.95\textwidth}{lCCC}
        \toprule
        Property & MM & Spider \cite{van2024spider} & SG \\ \midrule
        N subjects & 240 & 218 & 144 \\
        Sequences & CT & MRI (T1w, T2w) & CT, MRI Dixon (in-phase, fat, water) \\
        Spine Coverage & Whole spine & Lumbar only & Various parts\\
        Sex (\% female)               & 39 & 63 & 45 \\
        Age range (yrs)               & 31 -- 87 & 14 -- 84 & 20 -- 94\\
        Mean Age $\pm$ SD (yrs)              & $64 \pm 11$ & $60 \pm 15$ & $66 \pm 19$\\
        Is public & \xmark & \cmark & \xmark \\
        Date range & 2005 -- 2022 & 2019 -- 2020 & 2017 -- 2023\\
        Pathology & Multiple myeloma & Lower back pain & Diverse \\
        Manufacturers & Siemens, Philips & Siemens, Philips & Siemens, Philips, Canon, Toshiba, GE \\
        \bottomrule
    \end{tabularx}}
\end{table}

We use three datasets comprising seven imaging sequences: the public Spider lumbar spine MRI dataset \cite{van2024spider} and two private in-house datasets (SG and MM) spanning different modalities (CT, MRI), scanner manufacturers, imaging sequences, and spinal regions (cervical, thoracic, lumbar, sacral). Table \ref{tab:dataset_desc} shows more details about the datasets, and Figure \ref{fig:dataset-show} illustrates representative examples. Informed consent was waived by the local ethics committee (*blinded ID*). All datasets include manual reference segmentations from different expert raters. 

Datasets were preprocessed as follows: all reference segmentation masks were re-labeled to contain only three semantic classes: vertebrae, intervertebral discs, and spinal canal. Images were re-oriented to (Posterior, Inferior, Right)-order and resampled to a 1 mm$^3$ isotropic resolution. 

SG dataset was split into 80/10/10 subject-wise for training, validation, and testing, respectively. The Spider and MM datasets were used exclusively for evaluation. Training on Spider, which contains only lumbar scans, while the other datasets include a wider range of fields of view, introduces additional factors related to differences in spine coverage that would confound our analysis. The MM dataset is much more diverse, with subjects having multiple CT images from different scanners and noise levels, enabling more sophisticated out-of-distribution evaluation. Thus, we refrain from training on them and instead create a random 10/90 validation/test split for both datasets. The validation set for the latter datasets was used during development.

\subsection{Data Augmentation}%

\begin{figure}[!ht]
    \centering
    \includegraphics[width=0.99\linewidth]{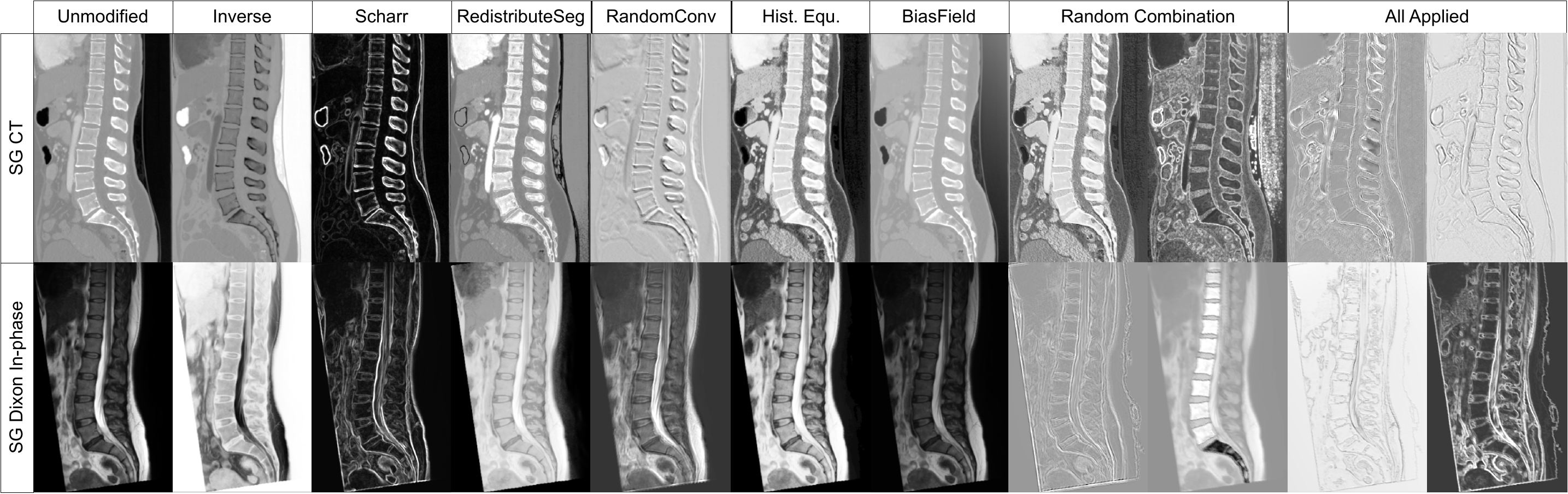}
    \caption{Example transforms applied to one SG CT (top) and SG Dixon in-phase image (bottom) of the same subject. The columns are the different transforms applied to these images.}
    \label{fig:transforms}
\end{figure}

We use the nnUNet framework \cite{isensee2021nnu}, a widely adopted framework for medical image segmentation. To address concerns that extensive data augmentation may slow training and hinder adoption, we developed a custom GPU nnU-Net. This trainer reproduces the default nnU-Net augmentation pipeline while extending it with additional GPU-based transformations that modify image appearance and simulate variations in imaging modality (CT vs. MRI) and contrast (e.g., T1- vs. T2-weighted MRI). The selected transformations are described in Table \ref{tab:augmentation_desc} and shown in the Figure \ref{fig:transforms}.

\begin{table}[!htb]
    \centering
    \caption{Summary of 3D data augmentation transforms used in this study.}
    \label{tab:augmentation_desc}
    {\fontsize{8}{9}\selectfont
    \begin{tabularx}{0.99\textwidth}{lX}
        \toprule
        Transform & Description \\ \midrule
        Intensity inversion \cite{hantze2025mri} & Min-Max intensity range inversion \\
        Scharr filtering \cite{saydazimov2025filter} & Gradient magnitude enhancement for edge emphasis \\
        RedistributeSeg \cite{warszawer2025totalspineseg} & Segmentation-driven regional intensity redistribution by adding a randomly scaled probability density function of voxel intensities within each segmented region. \\
        RandomConv \cite{xu2020robust} & Random 3D convolutions to induce texture variation \\
        Histogram equalization \cite{warszawer2025totalspineseg} & Estimate the intensity histogram, compute its normalized cumulative distribution function, and use it to remap voxel intensities. \\
        Bias field \cite{tustison2010n4itk} & Simulating intensity shading inhomogeneity \\
        Unsharp masking \cite{deng2010generalized} & High-frequency reinforcement for contour sharpening \\
        Function transform & Apply a randomly selected nonlinear function \\
        \bottomrule
    \end{tabularx}}
\end{table}%

Standard geometric transformations (e.g., rotations and flips) and other baseline augmentations (e.g., Gaussian noise, blurring, and resampling) are kept with the same configuration as the normal nnUNet trainer. The order of the transforms is set to geometric transformations first, then the new augmentations described in Table \ref{tab:augmentation_desc}, and finally the remaining default augmentations from the normal nnUNet trainer. We applied each augmentation with specific parameters and probabilities (for the exact configuration, see \href{https://osf.io/7y495/overview?view_only=6ad2940fe12346fdb528bfa871c82e0c}{blinded hyperlink}).

\subsection{Experiments}

Models were trained using a single sequence from the SG dataset. Validation only used in-sequence samples, whereas evaluation was performed on the test splits of all sequences/modalities described in the Dataset section.
We set the batch size to 2 and the patch size to (128, 128, 128). We trained for 1,000 epochs with the default nnUNetPlans, corresponding to the PlainConvUNet architecture \cite{isensee2021nnu}.

Our custom trainer was compared  under the same data setting against the standard nnUNetTrainer baseline and the official nnUNetTrainerDA5 configuration, which represents the most extensive default nnUNet augmentation strategy. All models were evaluated quantitatively on the corresponding test sets for each sequence. 

To quantify the contribution of individual augmentations, we performed ablation experiments in which each newly added augmentation was enabled with an application probability of 0.5, while disabling all other new augmentations. Additionally, disabling all of the baseline transformations highlights the difference our new augmentations make. Finally, we trained once while shuffling the order in which all but the geometric transformations are applied.

Segmentation performance was measured using the Dice similarity coefficient (per class).
Metrics were computed with Panoptica \cite{kofler2023panoptica}; the configuration we used for metric computation is available at \href{https://osf.io/7y495/overview?view_only=6ad2940fe12346fdb528bfa871c82e0c}{blinded hyperlink}. We recorded the training time using nnUNet’s internal timing and we averaged it across runs to compare the computational cost of our proposed GPU augmentation pipeline against the conventional nnUNet trainer. 

We assessed the statistical significance using the Wilcoxon signed-rank test; $p<0.05$ was considered significant.

For software and their versions, we use Python 3.11, \href{https://github.com/MIC-DKFZ/nnUNet}{nnUNetv2} v2.6.2, and \href{https://github.com/BrainLesion/panoptica}{panoptica} v1.1.4. We compute the Wilcoxon signed-rank with \href{https://github.com/scipy/scipy}{SciPy} v1.14.1. We managed the data using \href{https://github.com/Hendrik-code/TPTBox}{TPTBox} v0.4.2. Our GPU implementation heavily leans on \href{https://github.com/kornia/kornia}{kornia} v0.8.2.

\section{Results}

\begin{table}[!b]
    \centering
    \caption{Three training setups (rows) evaluated on the test splits across all seven sequences (columns) of the datasets, comparing the common nnUNet trainers with ours. Values are the global dice score averaged across classes, then averaged across subjects. Bold indicates the best-performing value across trainers. Dark brown values are in-domain evaluations. An asterisk indicates statistical significance against the best other setup.}
    \label{tab:results_}
    {\fontsize{8}{9}\selectfont
    \begin{tabularx}{0.99\textwidth}{lCCCCCCCC}
        \toprule
        && MM & \multicolumn{2}{c}{Spider} & \multicolumn{4}{c}{SG} \\ \cmidrule(lr){3-3}\cmidrule(lr){4-5}\cmidrule(l){6-9}
       Trained on & Setup & CT & T1w & T2w & CT & fat & in-phase & water \\ \midrule
SG in-phase & Base & $0.13\ast $& $0.76\ast $& $\textbf{0.83}$& $0.17\ast $& $0.79\ast $& $\ind{\textbf{0.91}}$& $0.82\ast $\\
SG in-phase & DA5 & $0.05\ast $& $0.77\ast $& $\textbf{0.83}$& $0.12\ast $& $0.82\ast $& $\ind{\textbf{0.91}}$& $0.82\ast $\\
SG in-phase & Ours & $\textbf{0.83}\ast $& $\textbf{0.82}\ast $& $\textbf{0.83}$& $\textbf{0.83}\ast $& $\textbf{0.87}\ast $& $\ind{0.90}\ast $& $\textbf{0.86}\ast $\\
\midrule
SG CT & Base & $\textbf{0.91}$& $0.15\ast $& $0.05\ast $& $\ind{\textbf{0.91}}$& $0.33\ast $& $0.15\ast $& $0.05\ast $\\
SG CT & DA5 & $\textbf{0.91}$& $0.26\ast $& $0.12\ast $& $\ind{\textbf{0.91}}$& $0.51\ast $& $0.21\ast $& $0.13\ast $\\
SG CT & Ours & $0.90\ast $& $\textbf{0.79}\ast $& $\textbf{0.78}\ast $& $\ind{\textbf{0.91}}$& $\textbf{0.75}\ast $& $\textbf{0.77}\ast $& $\textbf{0.71}\ast $\\
\midrule
SG fat & Base & $0.31\ast $& $\textbf{0.81}$& $0.81$& $0.41\ast $& $\ind{\textbf{0.89}}\ast $& $0.87$& $0.60\ast $\\
SG fat & DA5 & $0.31\ast $& $\textbf{0.81}$& $0.81$& $0.45\ast $& $\ind{\textbf{0.89}}\ast $& $0.87$& $0.56\ast $\\
SG fat & Ours & $\textbf{0.83}\ast $& $\textbf{0.81}$& $\textbf{0.82}$& $\textbf{0.83}\ast $& $\ind{0.88}\ast $& $\textbf{0.88}$& $\textbf{0.84}\ast $\\
\bottomrule
    \end{tabularx}}
\end{table}%
\begin{figure}[!b]
    \centering
    \includegraphics[width=0.95\linewidth]{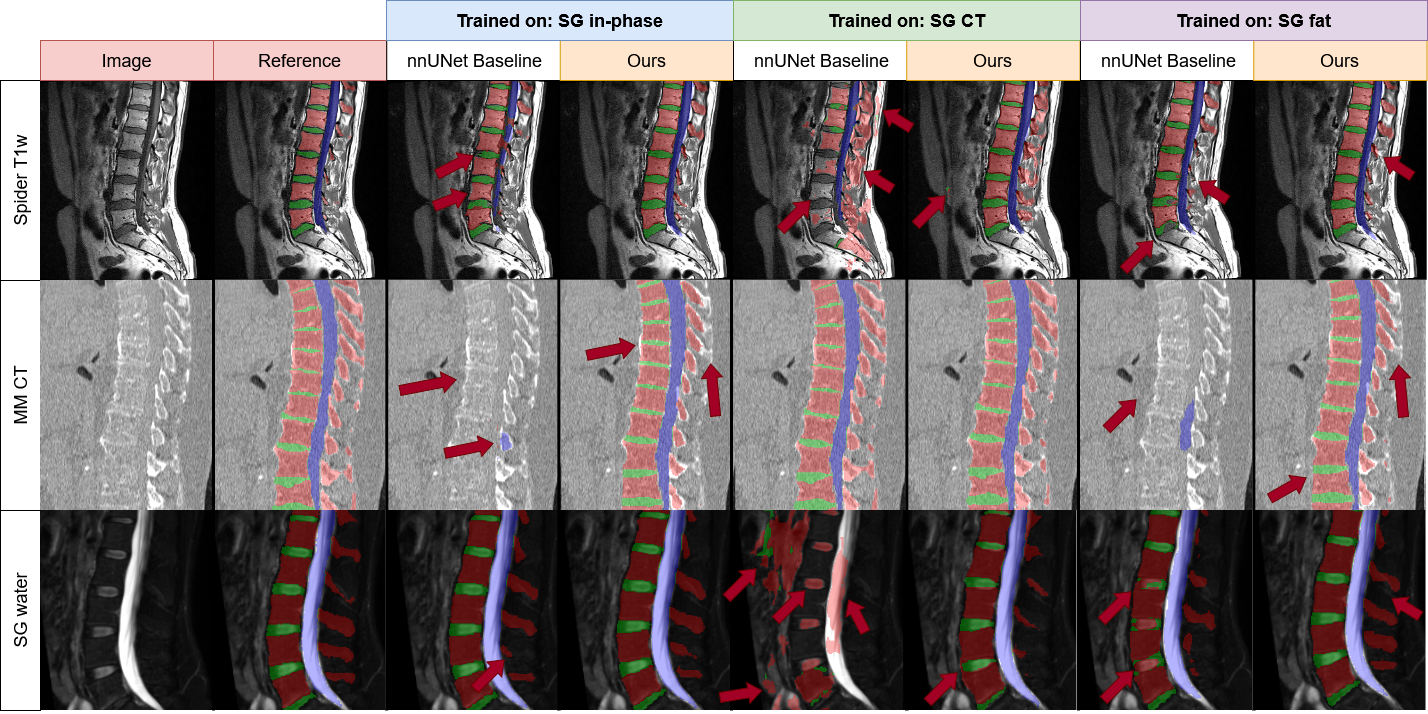}
    \caption{A qualitative comparison between predictions produced by the baseline and our setup on random test samples. The rows represent different samples, labeled with the dataset name and sequence. The columns display the ground-truth segmentation of the test image, along with the predicted segmentations from both the baseline and our method, for each of the three training sets. Red arrows highlight areas with prediction errors.}
    \label{fig:predictions}
\end{figure}%

\begin{table}[!htbp]
    \centering
    \caption{Class-wise global Dice score differences between three training setups comparing our method with the baseline nnUNetTrainer. Blue indicates performance gains, red indicates decreases; asterisks denote statistical significance.}
    \label{tab:results_diff}
    {\fontsize{8}{9}\selectfont
    \begin{tabularx}{0.99\textwidth}{lCCCCCCCC}
        \toprule
        && MM & \multicolumn{2}{c}{Spider} & \multicolumn{4}{c}{SG} \\ \cmidrule(lr){3-3}\cmidrule(lr){4-5}\cmidrule(l){6-9}
       Trained on & Class & CT & T1w & T2w & CT & fat & in-phase & water \\ \midrule
SG in-phase & Vertebra & $\posv{+0.70}\ast$ & $0.00\ast$ & $\negv{-0.01}\ast$ & $\posv{+0.59}\ast$ & $\posv{+0.02}\ast$ & $\negv{-0.01}\ast$ & $\posv{+0.04}\ast$ \\
SG in-phase & IVD & $\posv{+0.70}\ast$ & $\posv{+0.01}\ast$ & $0.00$ & $\posv{+0.60}\ast$ & $0.00$ & $\negv{-0.01}\ast$ & $\posv{+0.09}\ast$ \\
SG in-phase & Canal & $\posv{+0.70}\ast$ & $\posv{+0.17}\ast$ & $0.00$ & $\posv{+0.79}\ast$ & $\posv{+0.22}\ast$ & $\negv{-0.01}\ast$ & $\negv{-0.01}\ast$ \\ \midrule
SG CT & Vertebra & $\negv{-0.01}\ast$ & $\posv{+0.62}\ast$ & $\posv{+0.64}\ast$ & $\negv{-0.01}\ast$ & $\posv{+0.43}\ast$ & $\posv{+0.50}\ast$ & $\posv{+0.59}\ast$ \\
SG CT & IVD & $\negv{-0.03}\ast$ & $\posv{+0.64}\ast$ & $\posv{+0.69}\ast$ & $0.00\ast$ & $\posv{+0.52}\ast$ & $\posv{+0.52}\ast$ & $\posv{+0.57}\ast$ \\
SG CT & Canal & $0.00$ & $\posv{+0.64}\ast$ & $\posv{+0.86}\ast$ & $0.00$ & $\posv{+0.32}\ast$ & $\posv{+0.83}\ast$ & $\posv{+0.82}\ast$ \\ \midrule
SG fat & Vertebra & $\posv{+0.69}\ast$ & $\negv{-0.01}\ast$ & $\negv{-0.01}\ast$ & $\posv{+0.57}\ast$ & $\negv{-0.01}\ast$ & $0.00$ & $\posv{+0.18}\ast$ \\
SG fat & IVD & $\posv{+0.68}\ast$ & $\posv{+0.02}\ast$ & $\posv{+0.01}$ & $\posv{+0.59}\ast$ & $\negv{-0.01}\ast$ & $\posv{+0.01}$ & $\posv{+0.39}\ast$ \\
SG fat & Canal & $\posv{+0.19}\ast$ & $\negv{-0.01}\ast$ & $\posv{+0.01}\ast$ & $\posv{+0.10}\ast$ & $\negv{-0.01}\ast$ & $\posv{+0.01}$ & $\posv{+0.14}\ast$ \\
\bottomrule
    \end{tabularx}}
\end{table}%

\subsection{Segmentation performance}
Table \ref{tab:results_} shows how our custom trainer compares against the two nnUNet setups, the nnUNetTrainer (Base) and nnUNetTrainerDA5 (DA5). Our augmentation setup shows superior Dice performance on out-of-domain sequences across all three training setups. We observe this performance gain not only across the three semantic classes on average, but also quite consistently within each of them (see Table \ref{tab:results_diff}). Notably, the best overall boosts are for CT-MRI transfers, in both directions. On average across the three benchmarks and datasets, we reach an out-of-domain CT Dice performance of $0.83$ when training on MRI. This is $\simsym91\%$ of the respective average in-domain CT performance. Respectively, when training on CT, we reach an overall out-of-domain MRI performance that is $\simsym85\%$ of the respective in-domain MRI Dice scores. For qualitative results, see Figure \ref{fig:predictions}.
For in-domain samples, our setup sometimes yields slightly lower Dice scores (avg. $-0.008$ across classes and training sets).
Notably, baselines trained on SG Dixon in-phase or fat images perform well on Spider T1w and T2w images. We hypothesize that this is due to lumbar scans being easier to segment than other areas of the spine, as well as to T2w images being the same contrast as Dixon in-phase images.

The results of our ablations are in Table \ref{tab:results_ablation}. They confirm that the proposed augmentations are the main contributors to the observed out-of-distribution gains, with all newly introduced transformations showing statistically significant effects ($p < 0.001$). 

\begin{table}[!t]
    \centering
    \caption{Ablation evaluation using the models trained on SG in-phase and SG CT images. The first column indicates the changes to our setup. The values are global dice scores averaged across classes, then across subjects. Dark brown values are in-domain evaluations. Bold values are the best single-applied transformations for each testing set.}
    \label{tab:results_ablation}
    {\fontsize{8}{9}\selectfont
    \begin{tabularx}{0.99\textwidth}{lCCCCCcC}
        \toprule
        & MM & \multicolumn{2}{c}{Spider} & \multicolumn{4}{c}{SG} \\ \cmidrule(lr){2-2}\cmidrule(lr){3-4}\cmidrule(l){5-8}
       Setup & CT & T1w & T2w & CT & fat & in-phase & water \\ \midrule
\multicolumn{8}{c}{Trained on: SG in-phase} \\ 
\midrule
Ours & $0.83$ & $0.82$ & $0.83$ & $0.83$ & $0.87$ & $\ind{0.90}$ & $0.86$ \\
Ours (Base disabled) & $0.77$ & $0.81$ & $0.83$ & $0.81$ & $0.87$ & $\ind{0.90}$ & $0.86$ \\
Ours (Random order) & $0.83$ & $0.81$ & $0.83$ & $0.83$ & $0.87$ & $\ind{0.90}$ & $0.86$ \\
Base & $0.13$ & $0.76$ & $0.83$ & $0.17$ & $0.79$ & $\ind{0.91}$ & $0.82$ \\
\gmidrule
Base + Intensity inversion & $0.52$ & $0.79$ & $\textbf{0.83}$ & $0.60$ & $0.82$ & $\textbf{\ind{0.91}}$ & $0.84$ \\
Base + Scharr filtering & $0.65$ & $0.81$ & $\textbf{0.83}$ & $0.62$ & $0.84$ & $\textbf{\ind{0.91}}$ & $0.85$ \\
Base + RedistributeSeg & $\textbf{0.82}$ & $\textbf{0.82}$ & $\textbf{0.83}$ & $\textbf{0.80}$ & $\textbf{0.86}$ & $\textbf{\ind{0.91}}$ & $\textbf{0.86}$ \\
Base + Random Conv & $0.72$ & $0.81$ & $\textbf{0.83}$ & $0.75$ & $\textbf{0.86}$ & $\textbf{\ind{0.91}}$ & $\textbf{0.86}$ \\
Base + Histogram equalization & $0.45$ & $0.78$ & $\textbf{0.83}$ & $0.43$ & $0.81$ & $\textbf{\ind{0.91}}$ & $0.83$ \\
Base + Bias field & $0.44$ & $0.77$ & $\textbf{0.83}$ & $0.41$ & $0.80$ & $\textbf{\ind{0.91}}$ & $0.83$ \\
Base + Unsharp masking & $0.36$ & $0.76$ & $\textbf{0.83}$ & $0.31$ & $0.78$ & $\textbf{\ind{0.91}}$ & $0.83$ \\
Base + Function transform & $0.40$ & $0.77$ & $\textbf{0.83}$ & $0.38$ & $0.81$ & $\textbf{\ind{0.91}}$ & $0.83$ \\
\midrule
\multicolumn{8}{c}{Trained on: SG CT} \\
\midrule
Ours & $0.90$ & $0.79$ & $0.78$ & $\ind{0.91}$ & $0.75$ & $0.77$ & $0.71$ \\
Ours (Base disabled) & $0.90$ & $0.77$ & $0.76$ & $\ind{0.91}$ & $0.75$ & $0.75$ & $0.52$ \\
Ours (Random order) & $0.90$ & $0.79$ & $0.78$ & $\ind{0.90}$ & $0.75$ & $0.77$ & $0.72$ \\
Base & $0.91$ & $0.15$ & $0.05$ & $\ind{0.91}$ & $0.33$ & $0.15$ & $0.05$ \\
\gmidrule
Base + Intensity inversion & $\textbf{0.91}$ & $0.57$ & $0.48$ & $\textbf{\ind{0.91}}$ & $0.49$ & $0.31$ & $0.09$ \\
Base + Scharr filtering & $0.90$ & $0.57$ & $0.35$ & $\textbf{\ind{0.91}}$ & $0.53$ & $0.25$ & $0.07$ \\
Base + RedistributeSeg & $\textbf{0.91}$ & $\textbf{0.75}$ & $\textbf{0.73}$ & $\textbf{\ind{0.91}}$ & $\textbf{0.67}$ & $0.57$ & $0.15$ \\
Base + Random Conv & $0.90$ & $0.72$ & $0.68$ & $\textbf{\ind{0.91}}$ & $0.65$ & $\textbf{0.68}$ & $\textbf{0.27}$ \\
Base + Histogram equalization & $\textbf{0.91}$ & $0.31$ & $0.15$ & $\textbf{\ind{0.91}}$ & $0.40$ & $0.21$ & $0.09$ \\
Base + Bias field & $\textbf{0.91}$ & $0.39$ & $0.19$ & $\textbf{\ind{0.91}}$ & $0.48$ & $0.23$ & $0.08$ \\
Base + Unsharp masking & $\textbf{0.91}$ & $0.38$ & $0.19$ & $\textbf{\ind{0.91}}$ & $0.43$ & $0.21$ & $0.07$ \\
Base + Function transform & $\textbf{0.91}$ & $0.36$ & $0.18$ & $\textbf{\ind{0.91}}$ & $0.39$ & $0.20$ & $0.09$ \\
\bottomrule
    \end{tabularx}}
\end{table}%

\subsection{Training speed}
All models were trained on an Nvidia A40 GPU. On average across the SG sequences, the default nnUNetTrainer baseline required $\simsym 52$ s per epoch ($\simsym 14.5$ h total), while the nnUNetTrainerDA5 setup required $\simsym 59$ s per epoch ($\simsym 16.4$ h total). Re-implementing the base setup with our GPU optimizations reduces this to $\simsym 43$ s per epoch ($\simsym 12$ h total), which corresponds to a $\simsym 17\%$ speedup. When using all our proposed augmentations, training time becomes $\simsym 47$ s per epoch ($\simsym 13$ h total), retaining a $\simsym 10\%$ speedup (saving $\simsym 1.5$ h). Inference time remains unchanged since the architecture is identical.

\section{Discussion}

Our results demonstrate that a carefully selected set of data augmentations significantly enhances cross-domain robustness at minimal cost. Unlike prior approaches \cite{billot2023synthseg,graf2023denoising}, our method requires no auxiliary training or specialized pipelines. Thanks to GPU optimization, the more complex augmentation scheme does not increase training time and causes negligible, if any, in-domain performance loss.

Among these, \textit{RedistributeSeg} yields the largest benefits for cross-modality and cross-contrast transfer by exploiting the segmentation mask to independently adjust intensities inside and outside the region of interest, thereby generating localized intensity shifts that better capture modality-specific variations. Texture-oriented transformations, such as \textit{Scharr} and \textit{Random Convolution}, further enhance generalization by increasing textural diversity and more closely approximating appearance changes induced by modality or contrast differences. Combining our multiple augmentations also shows strong synergistic effects. For models trained on SG CT and evaluated on SG water images, single-transformation ablations reach a maximum Dice score of 0.27, compared to 0.71 when all augmentations are combined.
Although different transformation orderings produce visually distinct samples, randomly shuffling the order does not lead to significant performance gains. 

Several limitations should be acknowledged. Segmentations across datasets were generated by different annotators, and MRI–contrast–dependent tissue boundaries introduce ambiguity in the reference labels, contributing to variability. Moreover, spinal canal boundaries are not directly visible in CT, rendering their annotation inherently probabilistic. Finally, our evaluation is limited to semantic spine segmentation, and the generalizability of these findings to other tasks remains to be established.

Future work will investigate optimized augmentation that further leverages the segmentation mask, and evaluation will be extended to additional segmentation tasks. To facilitate reproducibility and adoption, we release our GPU-based augmentation framework as open source (\href{https://anonymous.4open.science/r/blinded-repository/README.md}{blinded hyperlink}) with native support for nnUNet and MONAI. Such augmentations can improve model performance across heterogeneous clinical imaging settings while shifting manual annotation efforts toward rare pathologies that are difficult to synthesize, rather than simple intensity-based variations.

%
%
%
\bibliographystyle{splncs04}
\bibliography{bib}
\end{document}